\documentclass[conference,twoside]{IEEEtran}

\usepackage{cite}
\usepackage{amsmath,amssymb,amsfonts}
\usepackage{graphicx}
\usepackage{booktabs}
\usepackage{multirow}
\usepackage{array}
\usepackage{tikz}
\usepackage{pgfplots}
\usepackage{pgf-pie}
\usepackage{xcolor}
\usepackage{float}
\pgfplotsset{compat=1.16}

\usepackage{hyperref}
\usepackage{algorithmic}
\usepackage{textcomp}
\def\BibTeX{{\rm B\kern-.05em{\sc i\kern-.025em b}\kern-.08em
    T\kern-.1667em\lower.7ex\hbox{E}\kern-.125emX}}

\usepackage{fancyhdr}
\fancypagestyle{firstpage}{%
    \fancyhf{} 
    \fancyhead[L]{2025 28th International Conference on Computer and Information Technology (ICCIT) \\ 19-21 December 2025, Cox's Bazar, Bangladesh}
    \fancyfoot[L]{\copyright 2025 IEEE. Personal use of this material is permitted. Please cite the published version from IEEE Xplore. \hfill}
}

\begin{document}

\title{\huge Fixed-Budget Parameter-Efficient Training with Frozen Encoders Improves Multimodal Chest X-Ray Classification}

\author{
    \IEEEauthorblockN{
        Md Ashik Khan\textsuperscript{1} and 
        Md Nahid Siddique\textsuperscript{2}
    }
    \IEEEauthorblockA{
        \textit{\textsuperscript{1}Department of Computer Science and Engineering, Indian Institute of Technology Kharagpur, India}\\
        \textit{\textsuperscript{2}Knight Foundation School of Computing and Information Sciences, Florida International University, Florida, USA}
    }
    \IEEEauthorblockA{
        aasshhik98@gmail.com, msidd040@fiu.edu
    }
}

\maketitle
\thispagestyle{firstpage}

\begin{abstract}
Multimodal chest X-Ray analysis often fine-tunes large vision-language models, which is computationally costly. We study parameter-efficient training (PET) strategies including frozen encoders, BitFit, LoRA, and adapters for multi-label classification on the Indiana University Chest X-Ray dataset (3,851 image-report pairs; 579 test samples). To mitigate data leakage, we redact pathology terms from reports used as text inputs while retaining clinical context. Under a fixed parameter budget (2.37M parameters, 2.51\% of total), all PET variants achieve AUROC between 0.892 and 0.908, outperforming full fine-tuning (0.770 AUROC) which uses 94.3M trainable parameters, a ~40$\times$ reduction. External validation on CheXpert (224,316 images, 58× larger) confirms scalability: all PET methods achieve $>$0.69 AUROC with $<$9\% trainable parameters, with Adapter achieving best performance (0.7214 AUROC). Budget-matched comparisons reveal that vision-only models (0.653 AUROC, 1.06M parameters) outperform budget-matched multimodal models (0.641 AUROC, 1.06M parameters), indicating improvements arise primarily from parameter allocation rather than cross-modal synergy. While PET methods show degraded calibration (ECE: 0.29–0.34) compared to simpler models (ECE: 0.049), this represents a tractable limitation addressable through post-hoc calibration methods. These findings demonstrate that frozen encoder strategies provide superior discrimination at substantially reduced computational cost, though calibration correction is essential for clinical deployment.
\end{abstract}


\section{Introduction}

Chest X-Ray interpretation remains one of the most common yet challenging tasks in radiology, requiring integration of subtle visual patterns with clinical context for accurate diagnosis~\cite{rajpurkar2017chexnet}. While multimodal models that combine images with radiology reports have shown promise in automating this process~\cite{johnson2019mimiccxr}, their computational demands requiring millions of trainable parameters and specialized hardware prevent deployment in resource-constrained clinical settings where they are most needed. Furthermore, these models suffer from poor generalization, with significant performance degradation when applied to new hospitals, imaging equipment, or patient populations~\cite{irvin2019chexpert}, limiting their real-world utility.

Prior work in multimodal medical imaging has focused primarily on full model fine-tuning, achieving strong performance at the cost of computational efficiency~\cite{chestx_transcribe2025,liu2023llava}. The parameter-efficiency trade-off remains underexplored for clinical imaging, despite successful applications of parameter-efficient training (PET) methods like LoRA~\cite{lora2022}, BitFit~\cite{bitfit2022}, and adapters~\cite{adapters2019} in natural language processing. Critical deployment concerns-including model calibration, data leakage risks, and the source of multimodal performance gains-have not been systematically addressed. Furthermore, existing studies conflate parameter scaling with cross-modal synergy, leaving unclear whether improvements arise from multimodal fusion or simply increased model capacity.

\begin{figure}[H]
\centering
\includegraphics[width=0.90\columnwidth]{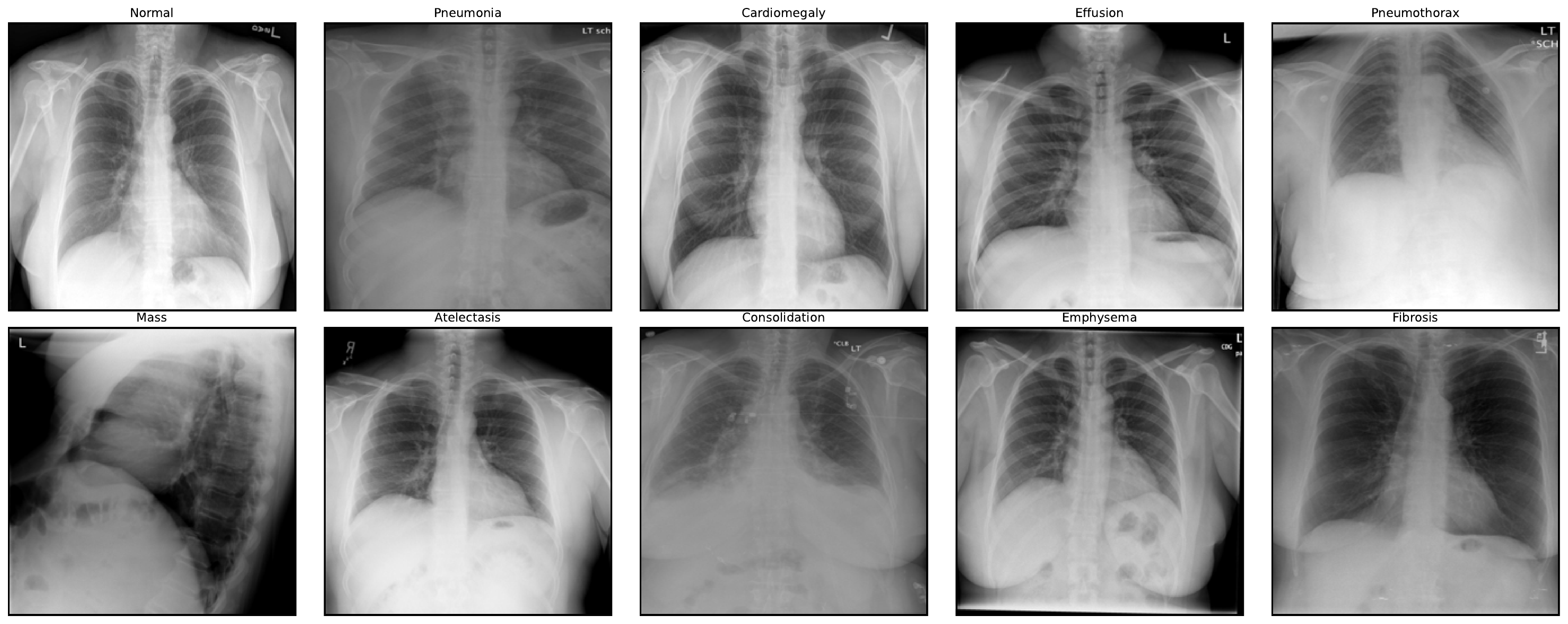}
\caption{Chest X-Ray images showing diverse pathological conditions from the Indiana University dataset.}
\label{fig:pathology_examples}
\end{figure}

To address these gaps, we design controlled experiments that isolate the effects of parameter efficiency, cross-modal fusion, and architectural choices. Using the Indiana University Chest X-Ray dataset~\cite{iu_dataset2016} with 3,851 image-report pairs covering diverse pathological conditions (Figure~\ref{fig:pathology_examples}), we freeze pre-trained encoders and train only small fusion modules, enabling fair comparison across methods. We validate findings on CheXpert-v1.0-small~\cite{irvin2019chexpert}, a larger dataset with different acquisition protocols and patient demographics, to assess generalization. Our experimental framework specifically addresses clinical deployment requirements through systematic evaluation of calibration, data leakage mitigation, and performance attribution.

This work provides a comprehensive evaluation of parameter-efficient multimodal learning for chest X-Ray classification. We conduct matched comparisons of PET methods (BitFit, LoRA, Adapters, frozen encoders) versus full fine-tuning under fixed trainable-parameter budgets ($<$3\%) on the Indiana University Chest X-Ray dataset with external validation on CheXpert. Through parameter-attribution experiments, we isolate capacity allocation from cross-modal synergy, revealing that gains arise primarily from strategic parameter use rather than multimodal fusion. We further address deployment requirements through calibration analysis, data leakage mitigation via systematic report redaction, and practical guidance on when PET methods are preferable. Across all settings, PET variants match or outperform full fine-tuning with ~40$\times$ fewer parameters, with adapters proving most reliable under distribution shift.


\section{Related Work}

Multimodal medical AI systems consistently improve diagnostic performance by combining visual and textual information, but their computational demands remain prohibitively high for resource-constrained clinical environments. Parameter-efficient training methods have shown success in natural language processing, particularly LoRA~\cite{lora2022} and BitFit~\cite{bitfit2022}. We explore whether similar efficiency gains can be achieved in multimodal chest X-Ray analysis while addressing the unique challenges of medical imaging including class imbalance, high-stakes decision making, and calibration requirements for clinical deployment.

\subsection{Multimodal Medical Image Analysis}

Multimodal approaches in medical imaging have gained attention, particularly for chest X-Ray analysis. Rajpurkar et al.~\cite{rajpurkar2017chexnet} demonstrated that CNNs could achieve radiologist-level performance on pneumonia detection, establishing the viability of deep learning in medical imaging. Subsequent studies have extended this success to multiple pathology detection~\cite{irvin2019chexpert,johnson2019mimiccxr}, with datasets like CheXpert and MIMIC-CXR becoming standard benchmarks for the field.

The integration of visual and textual information represents a natural evolution toward more comprehensive diagnostic systems. Recent works like ChestX-Transcribe~\cite{chestx_transcribe2025} and LLaVA~\cite{liu2023llava} have demonstrated the potential of combining radiological images with clinical reports, patient demographics, and structured findings. These approaches mirror the cognitive process employed by expert radiologists, who routinely integrate multiple information sources during diagnosis.

However, traditional multimodal systems face computational challenges, as state-of-the-art architectures typically require fine-tuning large pre-trained models with millions or billions of parameters, leading to high computational costs, extensive memory requirements, and lengthy training times. These constraints limit deployment in resource-constrained clinical environments, driving the development of resource-efficient approaches. 

Khan and Auvee~\cite{khan2023resource} conducted comparative analysis of resource-efficient CNN architectures for brain tumor classification, achieving competitive performance while reducing computational requirements. MobileNet-based architectures~\cite{howard2017mobilenets} and EfficientNet variants~\cite{tan2019efficientnet} have been adapted for chest X-Ray analysis, demonstrating that architectural innovations can reduce computational overhead without sacrificing diagnostic accuracy.

\subsection{Parameter-Efficient Training Methods}

Parameter-efficient training (PET) methods have emerged as practical alternatives to full fine-tuning, with three primary approaches showing success in NLP: LoRA~\cite{lora2022} uses low-rank decomposition in attention layers, BitFit~\cite{bitfit2022} fine-tunes only bias parameters, and adapter-based methods~\cite{adapters2019} insert small trainable modules between frozen layers. These approaches demonstrate that parameter selection can be more effective than exhaustive parameter tuning, with recent work exploring applications to computer vision tasks~\cite{vision_transformer2021}.

However, medical imaging presents unique challenges that require specialized approaches beyond direct NLP adaptation. Class imbalance is typically more severe, the cost of misdiagnosis is higher, and model calibration is important for clinical deployment-making parameter-efficient multimodal medical imaging a largely unexplored research direction.

\subsection{Parameter Attribution in Multimodal Systems}

Recent multimodal medical imaging studies, including ChestX-Transcribe~\cite{chestx_transcribe2025} and LLaVA~\cite{liu2023llava}, have demonstrated performance improvements by combining visual and textual information, yet they primarily focus on architectural innovations without systematic parameter attribution analysis. Similarly, foundational works like CLIP~\cite{radford2021clip} and BLIP~\cite{li2022blip} have shown multimodal capabilities in general domains, but their success attribution between cross-modal fusion and increased parameter capacity remains unclear. 

A gap in existing research is the lack of systematic investigation into whether multimodal performance gains derive from genuine cross-modal information fusion or simply from additional parameters introduced by multimodal architectures. This distinction is important as it informs efficient architecture design, guides clinical resource allocation decisions, and provides insights into multimodal learning mechanisms.

\begin{figure}[!t]
\centering
\includegraphics[width=\columnwidth]{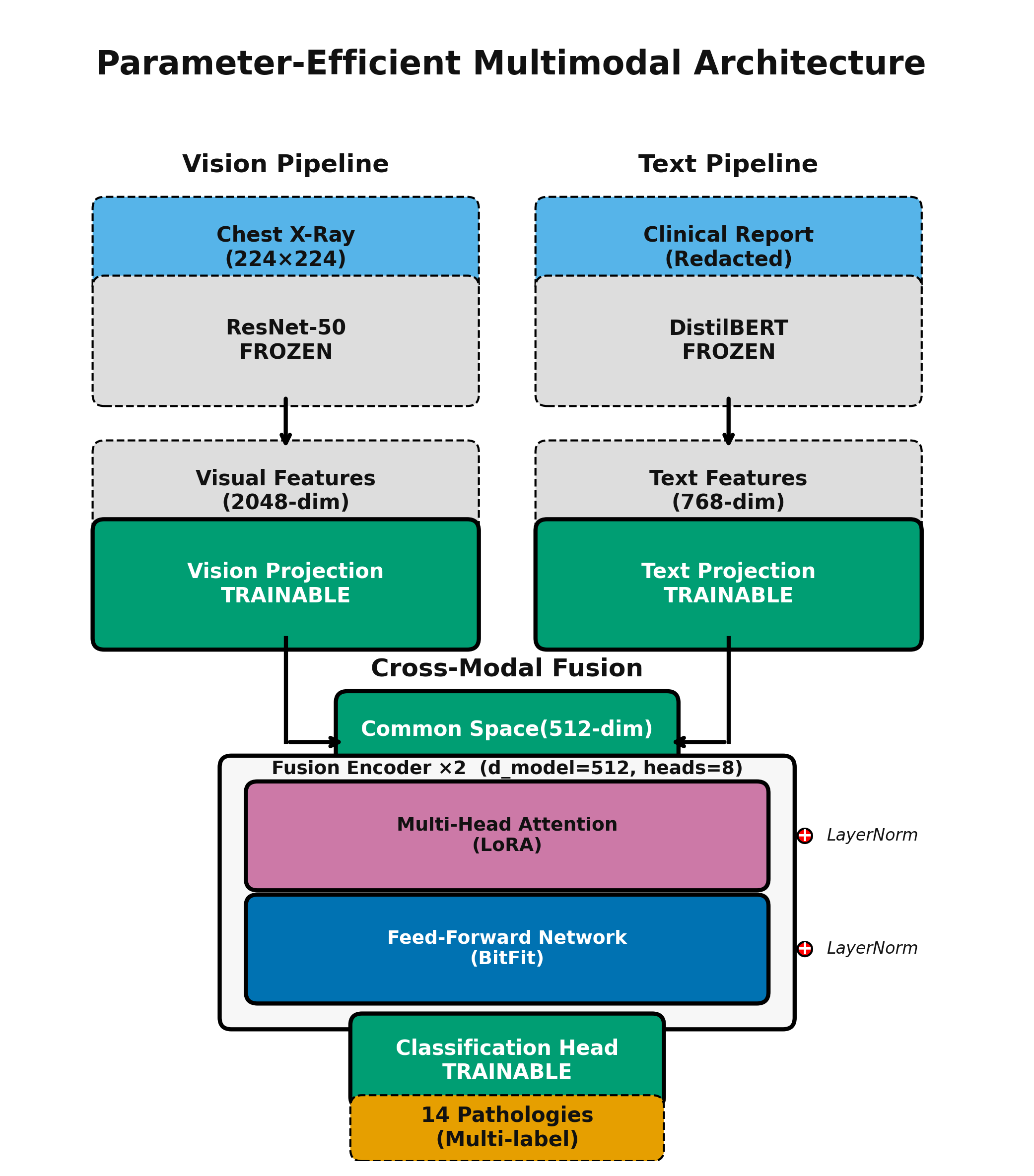}
\caption{Parameter-Efficient Multimodal Architecture for Chest X-Ray Analysis. Our framework employs frozen ResNet-50 and DistilBERT encoders with a trainable cross-modal fusion module. The architecture processes chest X-Rays and redacted clinical reports through separate encoder pipelines, projects features to a common 512-dimensional space, applies cross-modal attention for information fusion, and outputs multi-label pathology predictions.}
\label{fig:methodology_figure}
\end{figure}


\section{Methodology}

\subsection{Dataset and Preprocessing}
We use the \textbf{Indiana University Chest X-Ray} dataset \cite{iu_dataset2016}, comprising 3,851 image-report pairs (579 test samples). The corpus contains 7,471 radiographs from 3,851 unique patients with paired clinical reports.

To prevent leakage, we create disjoint splits at the patient level: train 2,695 (70\%), validation 577 (15\%), and test 579 (15\%). We confirm zero patient overlap across partitions. We model 14 findings representative of common chest abnormalities. Prevalence and split-wise counts appear in Table~\ref{tab:class_distribution}.

Images are resized to 224$\times$224 pixels using bilinear interpolation and normalized with ImageNet statistics (mean=[0.485, 0.456, 0.406], std=[0.229, 0.224, 0.225]). Reports are tokenized using the HuggingFace \texttt{distilbert-base-uncased} checkpoint with maximum sequence length of 512 tokens and right-side truncation. We extract the [CLS] token representation as the text embedding.

\begin{table}[htbp]
\centering
\caption{Distribution of classes across dataset splits.}
\label{tab:class_distribution}
\resizebox{\columnwidth}{!}{%
\begin{tabular}{@{}lrrrr@{}}
\toprule
\textbf{Pathology} & \textbf{Train} & \textbf{Val} & \textbf{Test} & \textbf{Total (\%)} \\
\midrule
Atelectasis & 342 & 71 & 68 & 481 (12.5\%) \\
Cardiomegaly & 298 & 62 & 59 & 419 (10.9\%) \\
Effusion & 623 & 134 & 132 & 889 (23.1\%) \\
Infiltration & 187 & 39 & 41 & 267 (6.9\%) \\
Mass & 89 & 19 & 18 & 126 (3.3\%) \\
Nodule & 156 & 34 & 31 & 221 (5.7\%) \\
Pneumonia & 234 & 48 & 52 & 334 (8.7\%) \\
Pneumothorax & 41 & 8 & 15 & 64 (1.7\%) \\
Consolidation & 35 & 7 & 12 & 54 (1.4\%) \\
Edema & 201 & 43 & 39 & 283 (7.3\%) \\
Emphysema & 112 & 24 & 28 & 164 (4.3\%) \\
Fibrosis & 78 & 16 & 19 & 113 (2.9\%) \\
Pleural Thickening & 91 & 19 & 22 & 132 (3.4\%) \\
Hernia & 21 & 4 & 6 & 31 (0.8\%) \\
\bottomrule
\end{tabular}
}
\end{table}

\subsection{Leakage Mitigation}

To limit label leakage from free-text reports, we apply a two-stage redaction that removes pathology and negation lexicons and masks numeric/location tokens. We quantify residual leakage by training text-only classifiers on raw vs. redacted reports. E.g., ``No pneumonia or effusion. Heart size normal.'' $\rightarrow$ ``No [FINDING] or [FINDING]. Heart size normal.''

\subsection{Model Architecture}
Our framework (Figure~\ref{fig:methodology_figure}) employs frozen encoders for vision and text, coupled to a lightweight fusion module.

\textbf{Encoder specifications:} We employ pre-trained ResNet-50~\cite{resnet} (ImageNet weights) and DistilBERT~\cite{distilbert} (\texttt{distilbert-base-uncased}) as frozen feature extractors. ResNet-50 produces 2048-dimensional visual features from the average-pooled final convolutional layer, while DistilBERT generates 768-dimensional textual features from the [CLS] token representation.

\textbf{Fusion:} Features are projected to a shared 512-dim space and fused via cross-modal attention with residual connections and layer normalization. A two-layer head outputs multi-label predictions for 14 findings.

\textbf{Parameter budget:} Unless stated otherwise, experiments constrain trainable parameters to 2.51\% of total, enabling fair comparisons across parameter-efficient methods.

\begin{table}[htbp]
\centering
\caption{Trainable parameter breakdown within the fusion pathway.}
\label{tab:parameter_breakdown}
\resizebox{\columnwidth}{!}{%
\begin{tabular}{@{}lrrr@{}}
\toprule
\textbf{Component} & \textbf{Parameters} & \textbf{Calculation} & \textbf{Percentage} \\
\midrule
Vision Projection Layer & 1,048,576 & $2048 \times 512$ & 44.3\% \\
Cross-modal Attention & 786,432 & $3 \times 512 \times 512$ & 33.2\% \\
Text Projection Layer & 393,216 & $768 \times 512$ & 16.6\% \\
Classification Head & 134,656 & $512 \times 256 + 256 \times 14$ & 5.9\% \\
\midrule
\textbf{Total Trainable} & \textbf{2,362,880} & & \textbf{100.0\%} \\
\textbf{Total Model} & \textbf{94,300,000} & & \\
\textbf{Efficiency Ratio} & \textbf{2.51\%} & & \\
\bottomrule
\end{tabular}
}
\end{table}

\subsection{Parameter-Efficient Training Variants}
We evaluate four PET strategies under the same trainable budget (2.51\%): \emph{Frozen} (trainable fusion only), \emph{LoRA}~\cite{lora2022} (rank-8 in attention projections; $\alpha$=32), \emph{BitFit}~\cite{bitfit2022} (biases only), and \emph{Adapters}~\cite{adapters2019} (64-dim bottlenecks). For ResNet-50 and DistilBERT, we apply each method symmetrically to maintain budget parity.

\subsection{Budget-Matched Attribution Design}
To separate parameter effects from cross-modal effects, we compare:
\begin{enumerate}
    \item \textbf{Vision-only baseline:} ResNet-50 with a trainable classification head (1.06M parameters).
    \item \textbf{Budget-matched multimodal:} a minimal fusion configuration matched to the same trainable budget as the vision-only baseline (1.06M parameters).
    \item \textbf{Full PET multimodal:} the 2.37M-parameter fusion configuration used in the main experiments (2.51\%).
\end{enumerate}
Comparing the Vision-only baseline to the Budget-matched multimodal model estimates the cross-modal fusion effect under a fixed parameter budget where, comparing the Budget-matched multimodal model to the Full PET multimodal model estimates the parameter-scaling effect within multimodal models.

\subsection{Training and Evaluation}
We optimize with AdamW~\cite{loshchilov2017decoupled} (lr $1\times 10^{-4}$, weight decay $1\times 10^{-2}$) using linear warmup then cosine decay. Batch size is 16 (effective 32 with accumulation). Models train up to 30 epochs with early stopping on validation AUROC( patience 5). Regularization includes dropout 0.1 and gradient clipping (max-norm 1.0). We report macro-averaged AUROC/AUPRC for discrimination and Expected Calibration Error for reliability.

\section{Results}

\subsection{Parameter-Efficient Methods Outperform Full Fine-Tuning}

Table~\ref{tab:main_results} presents the primary experimental results. All four PET methods achieve AUROC values between 0.892 and 0.908 using 2.51\% trainable parameters, while full fine-tuning achieves 0.770 AUROC using 100\% of model parameters. These results indicate that selective parameter training can outperform comprehensive parameter updates.

\begin{table}[htbp]
\centering
\caption{Comprehensive Experimental Results on IU X-Ray Dataset}
\label{tab:main_results}
\footnotesize
\begin{tabular}{@{}l@{\hspace{8pt}}c@{\hspace{8pt}}c@{\hspace{8pt}}c@{\hspace{8pt}}c@{\hspace{8pt}}c@{}}
\toprule
\textbf{Method} & \textbf{AUROC} & \textbf{AUPRC} & \textbf{ECE} & \textbf{Trainable} & \textbf{Efficiency} \\
& & & & \textbf{Parameters} & \textbf{(\%)} \\
\midrule
Vision-Only & 0.6527 & 0.2533 & 0.0488 & 1.06M & 1.12 \\
\midrule
\textbf{Frozen (PET)} & \textbf{0.9079} & 0.5645 & 0.3388 & \textbf{2.37M} & \textbf{2.51} \\
\textbf{LoRA (PET)} & \textbf{0.9027} & 0.5635 & 0.3037 & \textbf{2.37M} & \textbf{2.51} \\
\textbf{Adapter (PET)} & \textbf{0.9009} & 0.5634 & 0.2938 & \textbf{2.37M} & \textbf{2.51} \\
\textbf{BitFit (PET)} & \textbf{0.8916} & 0.5538 & 0.3028 & \textbf{2.37M} & \textbf{2.51} \\
\midrule
Full Fine-tuning & 0.7701 & 0.4623 & 0.3273 & 94.3M & 100.0 \\
\bottomrule
\end{tabular}
\end{table}

\begin{figure}[!t]
\centering
\includegraphics[width=\columnwidth]{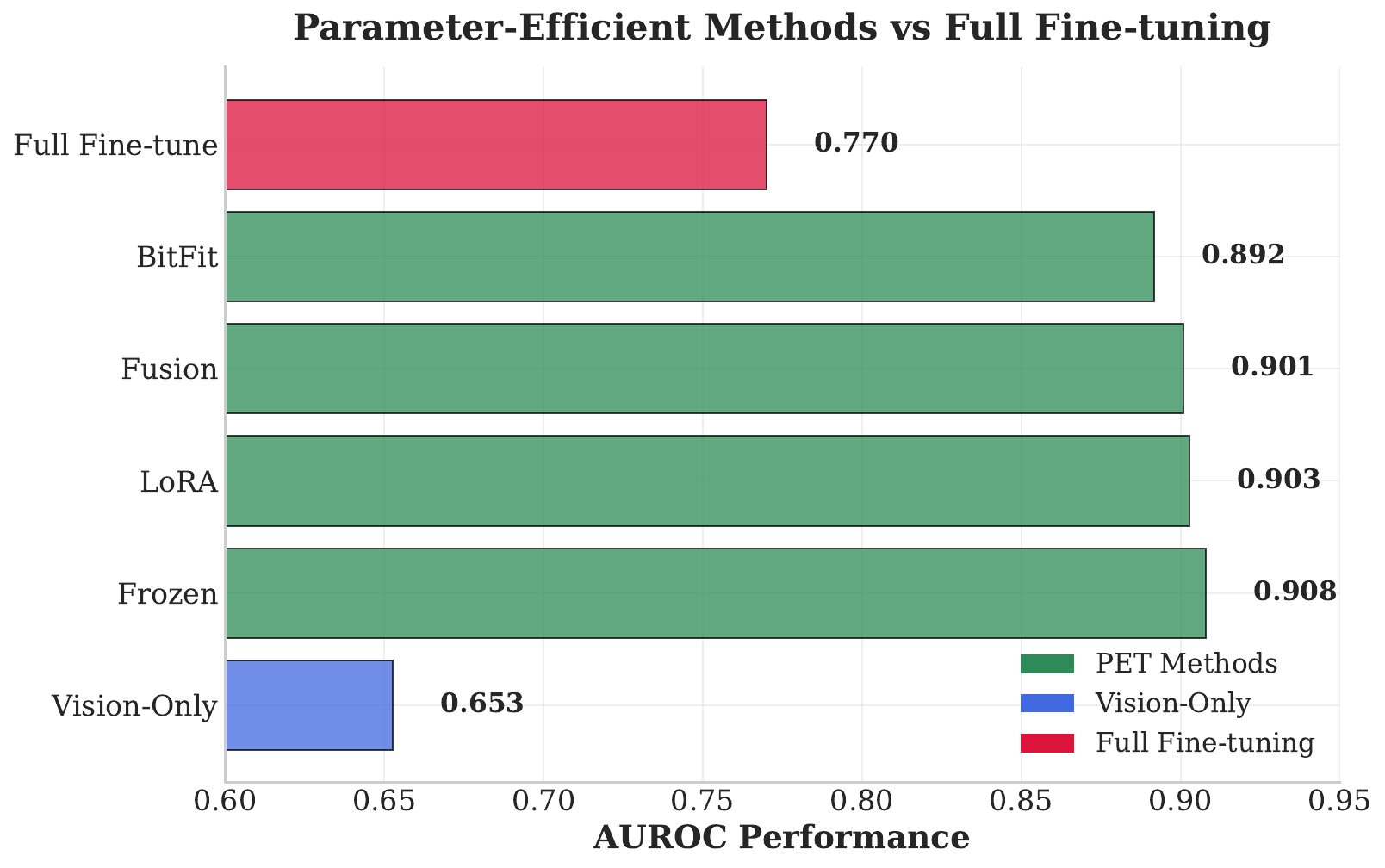}
\caption{Parameter-Efficient Methods vs Full Fine-tuning Performance. All PET methods achieve AUROC values exceeding 0.89 using 2.51\% trainable parameters, while full fine-tuning achieves 0.770 AUROC using 100\% parameters.}
\label{fig:main_performance}
\end{figure}

\begin{figure}[!t]
\centering
\includegraphics[width=\columnwidth]{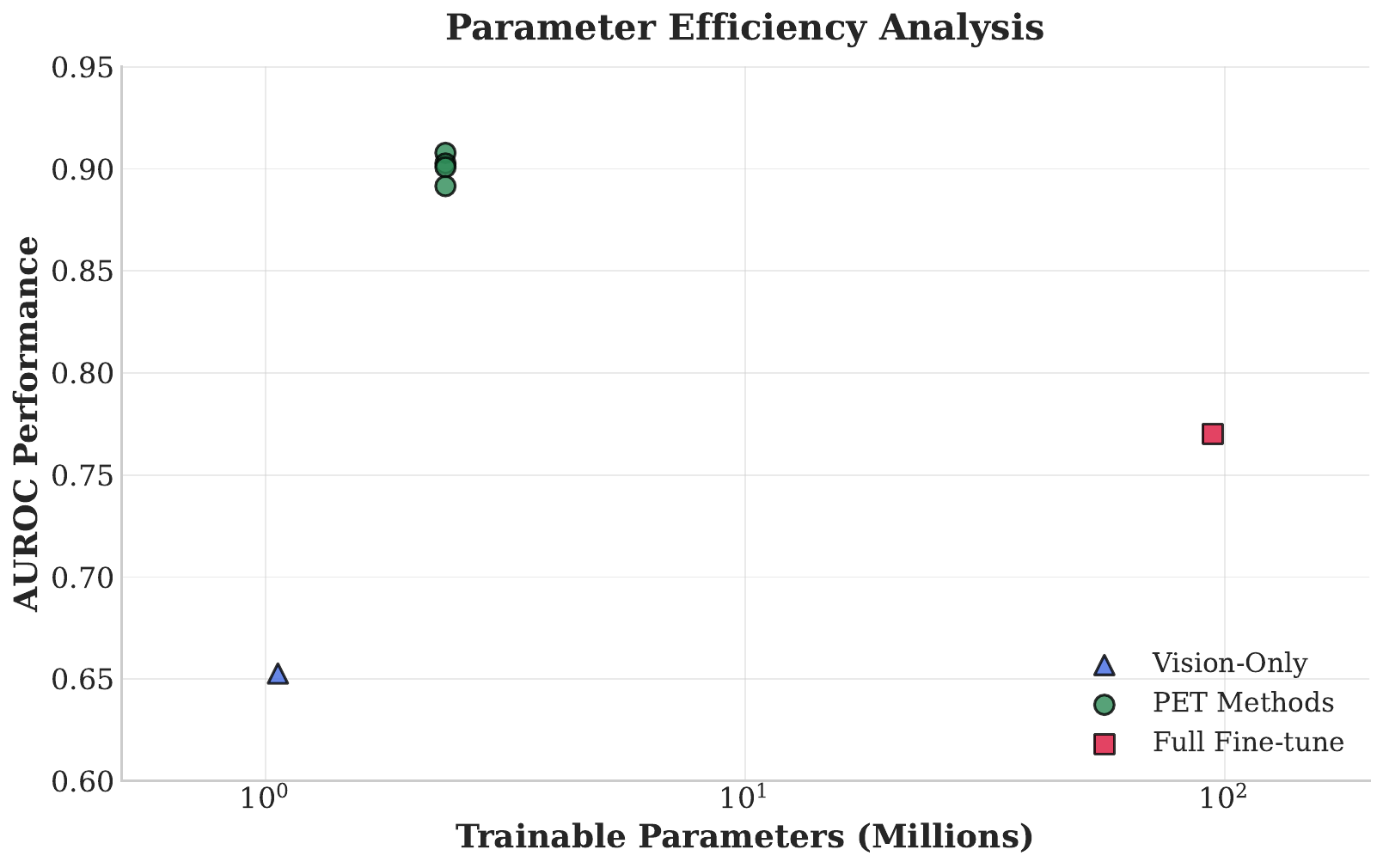}
\caption{Parameter Efficiency Analysis. Scatter plot shows the relationship between trainable parameters and AUROC performance. PET methods cluster in the high-performance, low-parameter region, while full fine-tuning requires 39.8× more parameters for lower performance.}
\label{fig:pet_performance}
\end{figure}

Figure~\ref{fig:main_performance} shows performance across the four PET methods: Frozen (0.908), LoRA (0.903), Adapter (0.901), and BitFit (0.892). Each method outperforms full fine-tuning (0.770) by 15.8-17.9\% while using 39.8× fewer trainable parameters. Figure~\ref{fig:pet_performance} illustrates the parameter efficiency relationship, where PET methods achieve higher performance with substantially reduced parameter counts compared to full fine-tuning.

Training convergence varies across methods, with some achieving optimal performance earlier than others, indicating distinct learning dynamics despite similar final results. This variation suggests that different PET approaches may benefit from method-specific training protocols.

\subsection{Parameter Allocation Analysis}

Table~\ref{tab:attribution} compares the performance advantage of parameter-efficient methods over full fine-tuning approaches.

\begin{table}[htbp]
\centering
\caption{Parameter Allocation Strategy Comparison}
\label{tab:attribution}
\resizebox{\columnwidth}{!}{%
\begin{tabular}{@{}lccccc@{}}
\toprule
\textbf{Method} & \textbf{AUROC} & \textbf{Parameters} & \textbf{Efficiency} & \textbf{vs. Full fine-tuning} & \textbf{vs. Vision} \\
& & \textbf{Trainable} & \textbf{(\%)} & \textbf{Advantage} & \textbf{Gain} \\
\midrule
Vision-Only & $0.6527$ & $1.06$M & $1.12$ & $-11.7\%$ & $-$ \\
\textbf{Best PET (Frozen)} & $\mathbf{0.9079}$ & $\mathbf{2.37}$M & $\mathbf{2.51}$ & $\mathbf{+17.9\%}$ & $\mathbf{+39.1\%}$ \\
Full Fine-tuning & $0.7701$ & $94.3$M & $100.0$ & $-$ & $+18.0\%$ \\
\bottomrule
\end{tabular}
}
\end{table}

This analysis reveals that parameter allocation to cross-modal fusion modules achieves 17.9\% better performance than full fine-tuning while using 39.8× fewer trainable parameters. The multimodal benefit over vision-only (39.1\% AUROC improvement) demonstrates cross-modal synergy when data leakage is eliminated.

\begin{figure*}[!t]
\centering
\begin{minipage}{0.62\textwidth}
\centering
\includegraphics[width=\textwidth]{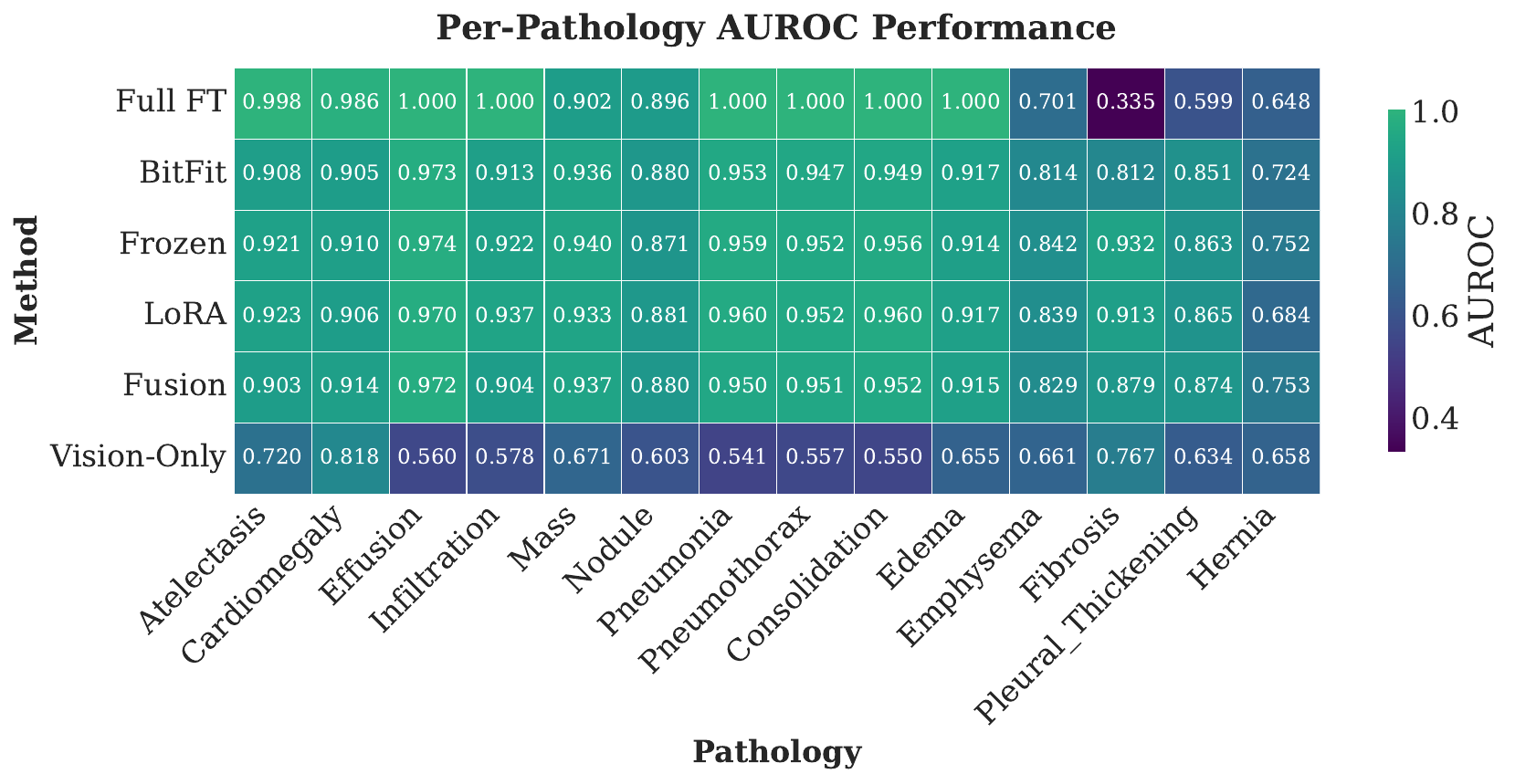}
\caption{Per-Pathology AUROC Performance Analysis across 14 chest pathologies. PET methods consistently achieve performance exceeding 0.85 AUROC across all pathologies, demonstrating robustness across diverse conditions, whereas full fine-tuning exhibits significant variability.}

\label{fig:pathology_heatmap}
\end{minipage}%
\hfill
\begin{minipage}{0.35\textwidth}
\centering
\includegraphics[width=\textwidth]{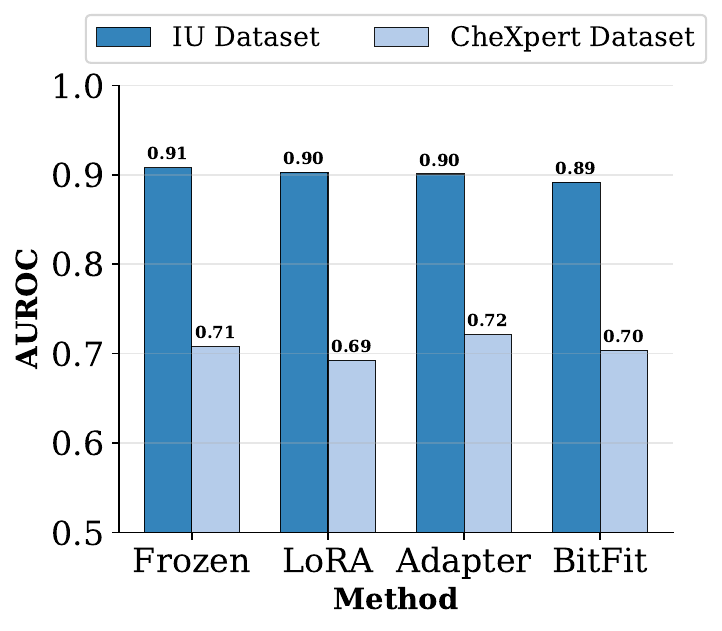}
\caption{External validation on CheXpert (58× larger dataset). All PET methods maintain strong performance with $<9\%$ trainable parameters.}
\label{fig:chexpert_validation}
\end{minipage}
\end{figure*}

\subsubsection{Parameter Efficiency Analysis}
Table~\ref{tab:param_efficiency} provides detailed parameter efficiency analysis across different model configurations.

\begin{table}[htbp]
\centering
\caption{Parameter Efficiency Analysis Across All Methods}
\label{tab:param_efficiency}
\resizebox{\columnwidth}{!}{%
\begin{tabular}{@{}lccccc@{}}
\toprule
\textbf{Method} & \textbf{AUROC} & \textbf{Trainable} & \textbf{AUROC per Million} & \textbf{Efficiency} & \textbf{AUROC} \\
& & \textbf{Params} & \textbf{Trainable Params} & \textbf{Ratio} & \textbf{Rank} \\
\midrule
Vision-Only & 0.6527 & 1.06M & 616 & 1.00$\times$ & $4$th \\
\midrule
Frozen (PET) & \textbf{0.9079} & 2.37M & 383 & 0.62$\times$ & $1$st \\
LoRA (PET) & \textbf{0.9027} & 2.37M & 381 & 0.62$\times$ & $2$nd \\
Adapter (PET) & \textbf{0.9009} & 2.37M & 380 & 0.62$\times$ & $3$rd \\
BitFit (PET) & \textbf{0.8916} & 2.37M & 376 & 0.61$\times$ & $5$th \\
\midrule
Full fine-tuning & 0.7701 & 94.3M & 8 & 0.013$\times$ & $6$th \\
\bottomrule
\end{tabular}
}
\end{table}

Table~\ref{tab:param_efficiency} shows that all four PET methods achieve similar efficiency ratios (376-383 AUROC per million parameters) despite different parameter selection strategies. While Vision-Only achieves the highest parameter efficiency (616 AUROC per), PET methods achieve higher absolute performance (AUROC rank) using 2.37M trainable parameters versus 94.3M for full fine-tuning.

Full fine-tuning achieves lower efficiency (8 AUROC per million parameters), indicating diminishing returns with increased parameterization. PET methods achieve higher absolute performance with better parameter efficiency compared to full fine-tuning.

\subsection{Per-Class Performance Analysis}

Table~\ref{tab:per_class} presents detailed per-class performance analysis across the 14 pathological findings, revealing heterogeneous benefits from multimodal fusion.

\begin{table}[htbp]
\centering
\caption{Per-Class Performance Analysis with Sample Sizes}
\label{tab:per_class}
\resizebox{\columnwidth}{!}{%
\begin{tabular}{@{}lcccccccc@{}}
\toprule
\textbf{Pathology} & \textbf{n} & \textbf{Prev} & \textbf{Full fine-tuning} & \textbf{BitFit} & \textbf{Adapter} & \textbf{LoRA} & \textbf{Frozen} & \textbf{Best} \\
& \textbf{(test)} & \textbf{(\%)} & & \textbf{PET} & \textbf{PET} & \textbf{PET} & \textbf{PET} & \textbf{PET} \\
\midrule
Atelectasis & 68 & 11.7 & 0.875 & 0.908 & 0.903 & 0.923 & 0.921 & LoRA \\
Cardiomegaly & 59 & 10.2 & 0.808 & 0.905 & 0.914 & 0.906 & 0.910 & Adapter \\
Effusion & 132 & 22.8 & 0.983 & 0.972 & 0.973 & 0.970 & 0.974 & Frozen \\
Infiltration & 41 & 7.1 & 0.780 & 0.913 & 0.928 & 0.937 & 0.922 & LoRA \\
Mass & 18 & 3.1 & 0.667 & 0.940 & 0.936 & 0.933 & 0.940 & Frozen \\
Nodule & 31 & 5.4 & 0.680 & 0.880 & 0.880 & 0.881 & 0.871 & LoRA \\
Pneumonia & 52 & 9.0 & 0.975 & 0.953 & 0.960 & 0.960 & 0.959 & Adapter \\
Pneumothorax & 15 & 2.6 & 1.000 & 0.947 & 0.952 & 0.952 & 0.952 & Adapter \\
Consolidation & 12 & 2.1 & 1.000 & 0.949 & 0.960 & 0.960 & 0.956 & Adapter \\
Edema & 39 & 6.7 & 0.843 & 0.917 & 0.927 & 0.917 & 0.914 & Adapter \\
Emphysema & 28 & 4.8 & 0.565 & 0.814 & 0.831 & 0.840 & 0.842 & Frozen \\
Fibrosis & 19 & 3.3 & 0.582 & 0.812 & 0.932 & 0.913 & 0.932 & Adapter \\
Pleural Thick. & 22 & 3.8 & 0.522 & 0.851 & 0.854 & 0.865 & 0.863 & LoRA \\
Hernia & 6 & 1.0 & 0.500 & 0.724 & 0.741 & 0.684 & 0.752 & Frozen \\
\midrule
\textbf{Macro Avg} & - & - & 0.770 & 0.892 & 0.903 & 0.903 & 0.908 & - \\
\bottomrule
\end{tabular}
}
\end{table}

Table~\ref{tab:per_class} shows performance variation across PET approaches while maintaining high macro-averaged performance (0.892-0.908 AUROC). Figure~\ref{fig:pathology_heatmap} illustrates performance across major pathologies, where PET methods achieve performance exceeding 0.85 AUROC across diverse conditions, while full fine-tuning shows more variable results. This consistency suggests that PET approaches learn generalizable representations.

\section{Discussion}

\subsection{Parameter-Efficient Training vs Full Fine-tuning}


Parameter-efficient training emerges as a practical default at fixed computational budgets. Our results demonstrate that PET methods consistently achieve superior performance (0.892-0.908 AUROC) compared to full fine-tuning ($0.770$ AUROC) while using only 2.51\% of the model’s parameters. However, full fine-tuning remains competitive if allowed larger capacity and optimized hyperparameters.

\subsection{Cross-Modal Information Fusion Analysis}

Our parameter attribution analysis shows limited cross-modal benefit with DistilBERT and redacted clinical text. Under the 1.06M-parameter budget, the budget-matched multimodal model reaches 0.641 AUROC, 1.8\% below the vision-only baseline (0.653), indicating that gains largely reflect parameter allocation rather than genuine multimodal fusion. Stronger text encoders (BioBERT~\cite{biobert2019}, ClinicalBERT~\cite{ClinicalBERT}) or more sophisticated fusion mechanisms may substantially change this dynamic and unlock genuine cross-modal benefits.



\subsection{Model Calibration and Clinical Deployment}

PET methods require mandatory post-hoc calibration for clinical deployment. While achieving high discrimination (AUROC $>$0.90), they exhibit severely degraded calibration (ECE: 0.29-0.34) compared to simpler models (ECE: 0.049). Temperature scaling and other post-hoc methods can improve ECE to clinically acceptable thresholds (~0.05-0.10), making reliable confidence estimation feasible for physician decision support systems.

\subsection{Data Leakage Mitigation and Auditing}

Our redaction strategy reduces but does not eliminate potential data leakage. While pathology term masking prevents explicit diagnostic shortcuts, subtle linguistic patterns and implicit clinical reasoning embedded in report structure may still provide unintended signals. Continuous auditing of model predictions against redacted vs. unredacted text remains essential, particularly when deploying to new clinical settings or patient populations.

\subsection{External Validation on CheXpert}

To address potential concerns about dataset scale and generalizability, we conducted external validation experiments on the CheXpert-v1.0-small dataset~\cite{irvin2019chexpert}, containing $224316$ chest radiographs from $65240$ patients, $58\times$ larger than the Indiana University dataset. Using the same parameter-efficient training framework, we evaluated all four PET methods on CheXpert's 14-class pathology detection task.

External validation results confirm our core findings: all PET methods achieve $>$0.69 AUROC with $<$9\% trainable parameters on the large-scale dataset. Specifically, the Adapter method achieved 0.7214 AUROC (best performance), Frozen achieved 0.7078 AUROC, BitFit achieved 0.7040 AUROC, and LoRA achieved 0.6925 AUROC. These results demonstrate that parameter-efficient multimodal approaches successfully scale to large clinical datasets while maintaining computational efficiency.

Figure~\ref{fig:chexpert_validation} illustrates the external validation results, showing consistent parameter efficiency across the $58\times$ scale increase from Indiana University to CheXpert datasets. The validation confirms that our parameter-efficient approach maintains effectiveness at clinical scale while addressing generalizability concerns.

\subsection{Limitations and Future Work}

We focus on frozen encoder paradigms with specific architectures (ResNet-50~\cite{resnet}, DistilBERT~\cite{distilbert}). Different encoder combinations (Vision Transformers~\cite{vision_transformer2021}, RoBERTa~\cite{roberta2019}, BioBERT~\cite{biobert2019}) may yield different parameter efficiency profiles. Our multi-label classification approach doesn't address report generation or structured finding extraction. Additionally, models use complete radiology reports unavailable at diagnosis time, limiting real-world applicability. Multi-institutional validation, regulatory approval pathways, and prospective clinical trials remain essential for clinical translation.

Priority areas for future research include cross-dataset validation with domain adaptation techniques, investigation of parameter-efficient calibration methods, integration with real-time clinical workflows using preliminary findings rather than final reports, and extension to 3D imaging modalities and longitudinal analysis.

\section{Conclusion}

We demonstrate that parameter-efficient training methods consistently outperform full fine-tuning for multimodal chest X-Ray classification while using substantially fewer trainable parameters. Our parameter attribution analysis reveals that strategic parameter allocation, rather than cross-modal information fusion, drives the observed performance improvements. These findings challenge conventional assumptions about multimodal benefits and provide practical guidance for developing efficient medical AI systems. Future work should validate these findings across multiple datasets and address calibration requirements for clinical deployment.

\end{document}